\documentclass{article}

\usepackage{arxiv}

\usepackage{forest}
\useforestlibrary{edges}
\usepackage{subcaption}
\usepackage{threeparttable}
\usepackage{amsmath}
\usepackage{lipsum}
\usepackage{multirow}
\usepackage{blindtext}
\usepackage{amssymb}
\usepackage{soul}
\usepackage{url}
\usepackage{geometry, color, graphicx, float}
\usepackage[colorlinks]{hyperref}
\usepackage{acronym}
\usepackage{natbib}
\usepackage{doi}
\usepackage{tikz}
\usepackage{amssymb}

\title{CLIPTime: Time-Aware Multimodal Representation Learning from Images and Text}

\author{ \href{https://orcid.org/0000-0000-0000-0000}{\includegraphics[scale=0.06]{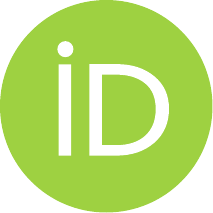}\hspace{1mm}Anju Rani}\thanks{Use footnote for providing further
		information about author (webpage, alternative
		address)---\emph{not} for acknowledging funding agencies.} \\
	Department of Energy Technology\\
	Aalborg University\\
	Esbjerg, Denmark 6700 \\
	\texttt{aran@energy.aau.dk} \\
	\And
	\href{https://orcid.org/0000-0000-0000-0000}{\includegraphics[scale=0.06]{orcid.pdf}\hspace{1mm}Daniel Ortiz-Arroyo} \\
	Department of Energy Technology\\
	Aalborg University\\
	Esbjerg, Denmark 6700 \\
	\texttt{doa@energy.aau.dk} \\
	\And
	\href{https://orcid.org/0000-0000-0000-0000}{\includegraphics[scale=0.06]{orcid.pdf}\hspace{1mm}Petar Durdevic} \\
	Department of Energy Technology\\
	Aalborg University\\
	Esbjerg, Denmark 6700 \\
	\texttt{pdl@energy.aau.dk} \\
}

\renewcommand{\shorttitle}{CLIPTime: Time-Aware Multimodal Representation Learning from Images and Text}

\hypersetup{
pdftitle={A template for the arxiv style},
pdfsubject={q-bio.NC, q-bio.QM},
pdfauthor={David S.~Hippocampus, Elias D.~Striatum},
pdfkeywords={First keyword, Second keyword, More},
}

\begin{document}
\maketitle

\begin{abstract}
Understanding the temporal dynamics of biological growth is critical across diverse fields such as microbiology, agriculture, and biodegradation research. Although vision-language models like Contrastive Language Image Pretraining (CLIP) have shown strong capabilities in joint visual-textual reasoning, their effectiveness in capturing temporal progression remains limited. To address this, we propose CLIPTime, a multimodal, multitask framework designed to predict both the developmental stage and the corresponding timestamp of fungal growth from image and text inputs. Built upon the CLIP architecture, our model learns joint visual-textual embeddings and enables time-aware inference without requiring explicit temporal input during testing. To facilitate training and evaluation, we introduce a synthetic fungal growth dataset annotated with aligned timestamps and categorical stage labels. CLIPTime jointly performs classification and regression, predicting discrete growth stages alongside continuous timestamps. We also propose custom evaluation metrics, including temporal accuracy and regression error, to assess the precision of time-aware predictions. Experimental results demonstrate that CLIPTime effectively models biological progression and produces interpretable, temporally grounded outputs, highlighting the potential of vision-language models in real-world biological monitoring applications.

The dataset and implementation are available in our open-source GitHub repository:
\href{https://github.com/PetarDurdevic/Funghi}{\texttt{github.com/PetarDurdevic/Funghi}}
\end{abstract}


\section{Introduction}
Vision-language models (VLMs) \cite{minaee2024large}, such as Contrastive Language–Image Pretraining (CLIP) \cite{radford2021learning}, have significantly advanced deep learning (DL) by unifying computer vision (CV) and natural language processing (NLP) through a shared multimodal embedding space. These models enable a broad range of tasks, including zero-shot classification and image-text retrieval, by encoding images and texts independently and aligning them in a common semantic space. Despite their strengths in spatial and semantic reasoning, a major limitation of current VLMs is their inability to natively process temporal information. This restricts their applicability in domains where understanding progression or sequential patterns is essential, such as biological process modeling \cite{sarch2024vlm}, \cite{cooper2025rethinking}, forecasting \cite{xue2025regression}, \cite{zhao2025images}, \cite{ouyang2018multi}, and time-sensitive monitoring applications \cite{hong2024cogvlm2}, \cite{mahmud2025integrating}. Recent efforts have aimed to extend VLMs by incorporating temporal dynamics, for example through temporal attention or sequence alignment in video-text tasks. Nevertheless, fully integrating spatial, semantic, and temporal reasoning within a unified vision-language framework remains a core challenge and a promising direction for future research.

Recent studies have begun to incorporate temporal reasoning into multimodal architectures. iTransformer \cite{liu2023itransformer} introduces a novel transformer-based model tailored for time-series forecasting. Unlike conventional transformers that apply attention across sequential time steps, iTransformer inverts this approach by applying attention over the feature dimension. This allows the model to capture inter-variable dependencies more effectively, which is particularly beneficial when handling multivariate time-series data. Similarly, PV-VLM \cite{lin2025pv} proposes a unified architecture for intra-hour photovoltaic (PV) power forecasting that integrates temporal, textual, and visual modalities. It comprises three specialized modules: a time-aware component based on PatchTST for modeling both local and global dependencies in PV time-series; a prompt-aware module that uses a large language model (LLM) to incorporate textual weather and domain-specific knowledge; and a vision-aware module that employs pretrained VLMs to extract semantic features from sky imagery. Expanding on this multimodal framework, Time-VLM \cite{zhong2025time} enhances time-series forecasting by converting temporal data into images, generating contextual textual descriptions, and integrating them via retrieval-augmented memory. This approach enriches multimodal reasoning and leads to notable gains in forecasting accuracy.

In parallel, several efforts have aimed to adapt CLIP-based architectures for temporal tasks, capitalizing on their strong vision-language grounding to enable time-aware reasoning. There is growing interest in extending CLIP to handle applications such as forecasting, temporal reasoning, and time-series analysis. Approaches like TC-CLIP \cite{kim2024leveraging}, C-TPT \cite{yoon2024c}, and others \cite{singh2024pattern}, \cite{he2024driver}, \cite{feng2023diverse} explore how temporal information can be integrated into vision-language models, either implicitly through visual progression or explicitly via temporal embeddings. These works underscore the potential of multimodal models to reason not only about what is observed, but also when it occurs. TC-CLIP \cite{kim2024leveraging} enhances video action recognition by incorporating temporal context into CLIP's architecture. Rather than analyzing each frame in isolation, it introduces specialized context tokens that aggregate information across frames, enabling the model to better understand temporal dependencies, particularly in low-label scenarios. C-TPT \cite{yoon2024c} improves the reliability and interpretability of CLIP’s predictions by modifying the text prompts through text-feature dispersion. This encourages greater diversity among prompts, which increases prediction robustness and enhances model transparency regarding uncertainty. Other approaches further demonstrate CLIP’s temporal versatility. For instance, \cite{singh2024pattern} converts time-series data into visual representations to leverage CLIP for retrieving visually similar patterns from a reference set, subsequently using the top-k matches for forecasting. In \cite{he2024driver}, CLIP is employed to predict driver intentions, such as turning or lane changes, by encoding both in-cabin and external video data alongside textual prompts. This method captures temporal dynamics like motion patterns and sequential behaviors, illustrating CLIP’s ability to model time-dependent phenomena and laying the groundwork for its application in time prediction tasks.

In this work, we extend the CLIP framework to reason jointly across visual, textual, and temporal modalities, aiming to predict both the stage of fungal growth and its associated timestamp. Fungal development follows a biologically ordered sequence, beginning with spores, transitioning through hyphae, and culminating in the formation of mycelium. To model this progression, we propose a multimodal architecture capable of capturing time-aware patterns without requiring explicit temporal input during inference. Our main contributions are summarized as follows:

\begin{enumerate}
\item We extend CLIP’s joint vision-language embedding space to support continuous time prediction, enabling the model to infer temporal dynamics from multimodal semantics.
\item We introduce a transformer-based temporal prediction head that employs self-attention to learn temporal dependencies within the fused CLIP embeddings.
\item We employ a multi-task learning framework that jointly performs fungal stage classification and timestamp regression, enhancing both representation learning and inference robustness.
\item We fine-tune the proposed model on a synthetically generated, time-annotated fungal growth dataset, showcasing CLIP’s adaptability to scientific domains that demand temporally structured predictions beyond typical vision-language tasks.
\end{enumerate}

The remainder of this paper is organized as follows: Section 2 presents the proposed methodology, including the creation of the synthetic fungal growth dataset and the design of the multimodal architecture for processing visual, textual, and temporal inputs. Section 3 describes the implementation details and reports the experimental results. Section 4 concludes the paper by summarizing the key findings and discussing their broader implications.

\begin{figure*}[htbp]
\centering
\includegraphics[width=\linewidth]{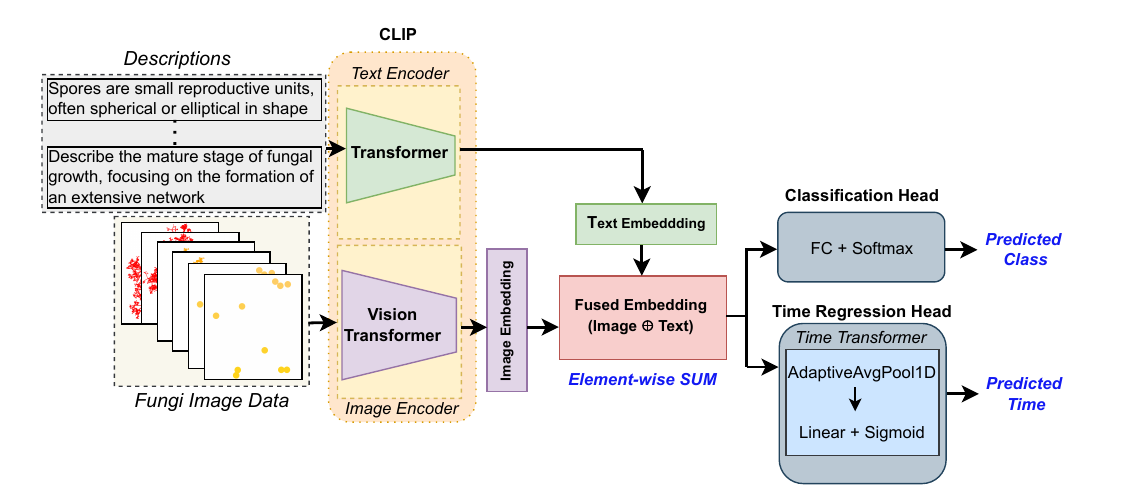}
\caption{Illustration of the proposed framework.}
\label{Fig1}
\end{figure*}

\section{Methodology}
This section describes the architecture of the proposed CLIPTime model, which is designed to jointly process and reason across multimodal inputs encompassing visual, textual, and temporal information. We begin by detailing our multi-task learning framework that simultaneously performs fungal growth stage classification and timestamp regression. This is followed by an overview of the synthetic fungal growth dataset used to train and evaluate the model.

\subsection{Multi-Task Fusion of CLIP}
Unlike traditional applications of CLIP that focus on zero-shot classification, CLIPTime repurposes the joint vision-language embedding space for time-aware inference. In this framework, images are first encoded using CLIP’s vision encoder, while corresponding textual prompts describing fungal stages are processed by the text encoder \cite{rani2025fungalzsl}. These embeddings are fused to form a shared multimodal representation that captures the semantic context of fungal development.

To perform multi-task prediction, two specialized heads are attached to this fused embedding. A simple fully connected (FC) layer is used to perform classification across the three fungal classes: spore, hyphae, and mycelium. Simultaneously, a transformer-based module, referred to as the Time-Transformer, is introduced for continuous timestamp regression. This module leverages self-attention mechanisms to capture temporal dependencies embedded within the multimodal features, enabling accurate prediction of growth progression without the need for explicit timestamp input during inference. The entire architecture is trained on a synthetically generated, time-aligned fungal dataset and fine-tuned using a multi-task learning objective that jointly optimizes classification and time regression error. This design highlights the adaptability of vision-language models to scientific domains requiring temporally structured reasoning, as illustrated in Figure~\ref{Fig1}.

\begin{figure*}[htbp]
\centering
\includegraphics[width=0.4\linewidth, angle=90]{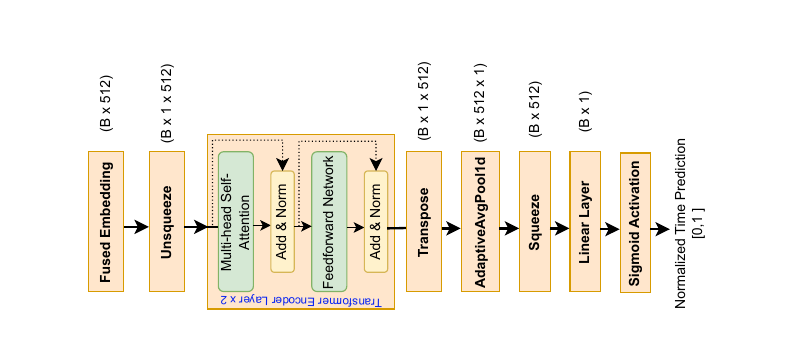}
\caption{Framework of the proposed Time Transformer.}
\label{Fig2}
\end{figure*}

\subsubsection{Time Normalization}
To ensure numerical stability during regression, the timestamps are linearly scaled to a normalized range of [0,1]. This normalization is performed using the following formula:

\begin{equation} \label{eq:1}
   t_i = \frac{t_i - t_{min}}{t_{max} - t_{min}}
\end{equation}
where, \(t_{min}\) and \(t_{max}\) represent the minimum and maximum timestamps in the dataset, respectively. These values are stored and used during inference to accurately re-project the normalized predictions back to the original real-time scale.

\begin{figure*}[htbp]
\centering
\includegraphics[width=\linewidth]{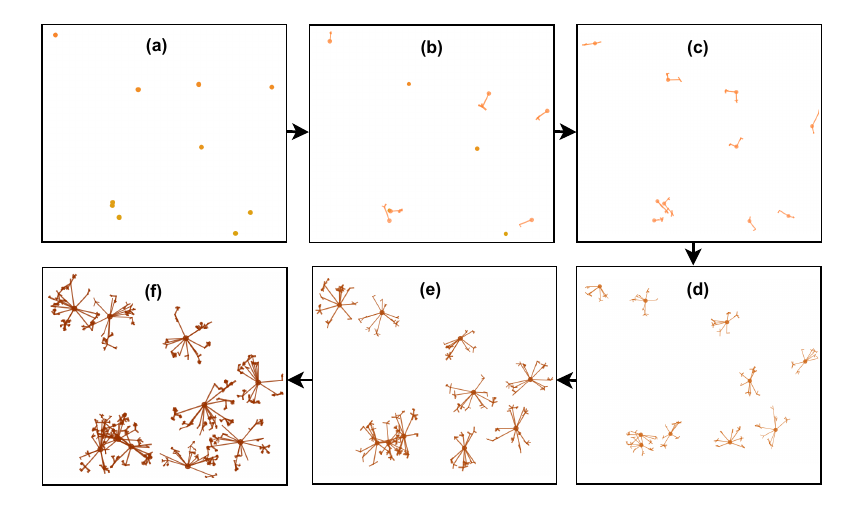}
\caption{Lifecycle of fungi growth: (a) Spore, (b)-(c) Spore to hyphae transition, (d) Hyphae, (e) Hyphae to mycelium transition, and (f) Mycelium.}
\label{Fig3}
\end{figure*}

\subsection{Time-Transformer}
The Time-Transformer is a lightweight, transformer-based regression head designed to predict a normalized timestamp \(\hat{t} \in [0, 1]\) from the fused image-text embedding generated by the CLIP backbone. Unlike traditional approaches that rely on explicit temporal tokens, this module captures implicit temporal progression by leveraging the rich visual and textual semantics embedded within CLIP’s joint representation space. Specifically, the time regression head operates on the fused embedding \(x \in \mathbb{R}^{d}\), which is obtained by element-wise summation of the image and text embeddings produced by CLIP. This combined embedding effectively encodes the evolving semantic patterns corresponding to different fungal growth stages. The detailed architecture of the Time-Transformer head is illustrated in Figure~\ref{Fig2}.

Let \(x \in \mathbb{R}^{B\times d}\) denote the fused embedding for a batch of size \(B\), where each embedding has a dimensionality \(d = 512\). To enable processing by the transformer module, this fused vector is reshaped to be treated as a sequence of length 1:

\begin{equation}
    x\in \mathbb{R}^{B\times d} \rightarrow X \in \mathbb{R}^{B\times 1\times d}
\end{equation}

Here, the sequence length is fixed at 1, indicating that each sample is represented by a single fused semantic unit combining both image and text embeddings. Due to this fixed-length sequence, no positional encoding is applied. This semantic representation is then fed through two stacked Transformer Encoder layers. The processing within each Transformer Encoder layer can be mathematically expressed as:

\begin{equation}
    Z_1 = \text{LayerNorm}(X + \text{MHSA}(X)),\quad \textrm{and} \quad
    Z_2 = \text{LayerNorm}(Z_1 + \text{FFN}(Z_1))
\end{equation}

where \( \text{MHSA}(\cdot) \) denotes multi-head self-attention, and \( \text{FFN}(\cdot) \) is a position-wise feed-forward network equipped with a ReLU activation function and a hidden layer dimension of 2048. After passing through the stacked encoder layers, the resulting output \(Z_{final} \in \mathbb{R}^{B \times 1 \times d}\) is aggregated using adaptive average pooling along the sequence length dimension, producing a fixed-size feature vector suitable for regression:

\begin{equation}
  z_{pooled} = AvgPool1D(Z_{final}) = Z_{final}\left [ :,0,: \right ] \in \mathbb{R}^{B\times d}
\end{equation}

This pooled vector is then passed through a fully connected linear layer, followed by a sigmoid activation to predict the normalized time value \(\hat{t}\), expressed as:

\begin{equation}
   \hat{t} = \sigma (z_{pooled}\ast w_t + b_t) \in [0,1]
\end{equation}

where \(w_t \in \mathbb{R}^{d}\) and \(b_t \in \mathbb{R}\) are the learnable parameters of the regression head, and \( \sigma(\cdot) \) represents the sigmoid activation, ensuring the output lies within the normalized range [0, 1].

\begin{equation}
t_{final} = \hat{t} \ast \left ( t_{max} - t_{min} \right ) + t_{min}
\end{equation}

Overall, the Time-Transformer introduces a novel approach to temporal regression by interpreting the fused semantics of visual and textual modalities as implicit indicators of temporal progression. By leveraging self-attention and feed-forward layers applied to this single embedding token, the module captures complex, non-linear relationships that map semantic evolution directly to real-valued timestamps. This design makes it well-suited for time-aware vision-language tasks where explicit temporal tokens or embeddings may be unavailable or undesirable. When integrated with the CLIPTime framework, the Time-Transformer significantly enhances the model’s ability to infer and align temporal stages with the underlying biological growth patterns captured in multimodal inputs.

\begin{figure}[htbp]
\centering
\includegraphics[width=0.7\linewidth]{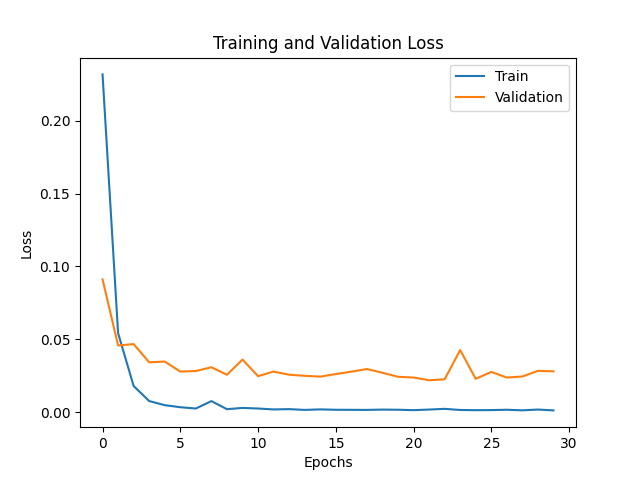}
\caption{Training and Validation Loss curve.}
\label{Fig4}
\end{figure}

\subsection{Loss Formulation}
The proposed framework utilizes a multi-task learning objective that jointly optimizes classification accuracy and temporal regression. Specifically, the model simultaneously predicts the fungal growth stage through a classification head and estimates the corresponding timestamp via the Time-Transformer regression head. By integrating these two complementary tasks, the framework better captures the complex biological dynamics of fungal development, resulting in improved representation learning and prediction performance.

\begin{itemize}
\item Classification Loss (\(\mathcal{L}_{class}\)): The classification loss is computed using the CrossEntropyLoss, which is standard for multi-class classification problems. Given predicted logits \(\hat{y} \in \mathbb{R}^{B\times C}\) and true class labels \(y \in ({1,..., C}^B)\), where \(B\) is the batch size and \(C\) the number of fungal growth classes, the loss measures the discrepancy between predicted and true distributions:
    
\begin{equation}
     \mathcal{L}_{class} = CE(\hat{y}, y)
\end{equation}
    
This function calculates the negative log-likelihood of the true class labels given the predicted probability distribution. It effectively encourages the model to assign higher confidence scores to the correct classes while penalizing incorrect predictions, thereby improving classification accuracy.
    
\item Time-regression Loss (\(\mathcal{L}_{time}\)): For the time regression task, the model outputs a scalar prediction \(\hat{t} \in [0, 1]\), representing the normalized estimated timestamp. This value is compared to the normalized ground truth time \(t \in [0, 1]\) using the Mean Squared Error (MSE) metric, which quantifies the average of the squared differences between predicted and true timestamps. Normalization is applied using the min-max scaling described in Equation~\ref{eq:1}, ensuring that all timestamps lie within the [0, 1] range. This normalization stabilizes the regression process and supports effective learning. The loss is computed as:
    
\begin{equation}
    \mathcal{L}_{time} = \frac{1}{B}\sum_{i=1}^{B}(\hat{t_i} - t_i)^2
\end{equation}
    
\end{itemize}

To guide the model in learning both discrete stage classification and continuous temporal progression, we define a composite objective function that integrates the two loss components:

\begin{equation}
    \mathcal{L}_{total} = \mathcal{L}_{cls} + \mathcal{L}_{time}
\end{equation}

This combined formulation ensures that the model jointly optimizes for accurate fungal stage recognition while grounding its predictions along a continuous temporal axis. To enable flexibility in balancing the contribution of each task, the loss function can be generalized with weighting coefficients \(\alpha\) and \(\beta\):

\begin{equation}
    \mathcal{L}_{total} = \alpha \ast \mathcal{L}_{cls} + \beta \ast \mathcal{L}_{time}
\end{equation}
In this study, we set both\(\alpha\) = 1 and \(\beta\) = 1, assigning equal importance to classification and regression during training.

\subsection{Time-Aligned Fungi Dataset}
We define the synthetic fungal growth dataset as \(D = \left\{ (x_i, l_i, t_i, d_i )\right\}_{i=1}^{N}\), where each data point comprises four components: an RGB image \(x_i \in \mathbb{R}^{H\times W\times 3}\), a categorical label \(l_i \in C\), a continuous timestamp \(t_i \in \mathbb{R}\), and a corresponding textual description\(d_i\). The image \(x_i\) visually captures the morphological features of fungal development, while the label \(l_i\) categorizes the sample into one of the three growth stages: spore, hyphae, or mycelium. The dataset spans a continuous temporal range from \(t_{min}\) to \(t_{max}\), with each image tagged by a timestamp \(t_i\) that reflects its position along the fungal lifecycle. This temporal annotation enables time-aware learning and regression. To further enhance semantic representation and facilitate multimodal reasoning, each sample is also accompanied by a natural language description \(d_i\), which provides contextual description about the fungal stage and its progression. The overall lifecycle and corresponding sample images are illustrated in Figure~\ref{Fig3}.

The three classes in the dataset reflect biologically significant stages of fungal development \cite{deacon2005fungal}, \cite{lutzoni2004assembling}, \cite{guarro1999developments}, each corresponding to a distinct phase in the fungal lifecycle. The initial stage, spore, represents the unicellular reproductive unit responsible for initiating growth \cite{lutzoni2004assembling}. Under favorable environmental conditions, spores germinate by producing a germ tube \cite{corona2023assessment}, which elongates into hyphae, long, filamentous structures that mark the onset of vegetative development \cite{bago1998architecture}. These hyphae branch and interconnect to form a growing network, facilitating substrate colonization and nutrient absorption \cite{boonruang2023study}, \cite{sephton2018pathways}. The final stage, mycelium, emerges as a dense, interconnected mesh of hyphae. It constitutes the mature fungal body, critical for resource acquisition, substrate decomposition, and, in many species, the formation of reproductive structures. Mycelium enables rapid territorial expansion and plays a key ecological role in sustaining fungal populations \cite{balmant2015model}, \cite{cano2019neurospora}. To facilitate reproducible, time-aware fungal analysis, this work introduces a synthetically generated, time-aligned fungal growth dataset, made publicly available for research use \cite{rani2025synthetic}. The dataset is analytically constructed to simulate clear distinctions between the developmental stages, aligning visual, textual, and temporal modalities. A detailed description of the dataset generation methodology is provided in \cite{FungalZSL}.

\begin{figure}[htbp]
\centering
\includegraphics[width=0.7\linewidth]{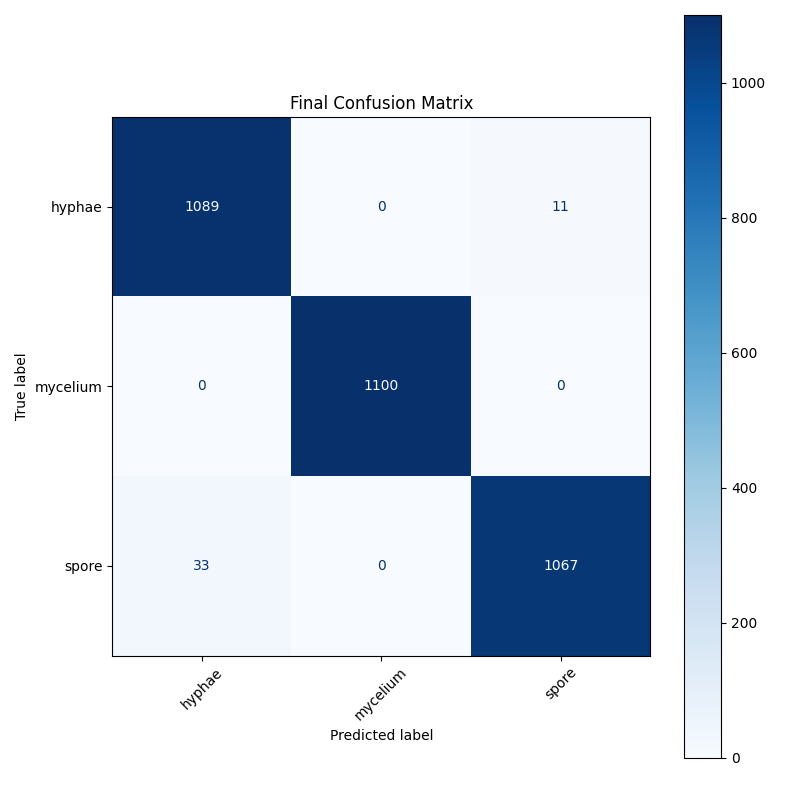}
\caption{Confusion matrix for the classification of hyphae, mycelium, and spore stages on the test data.}
\label{Fig5}
\end{figure}

\begin{figure}[htbp]
\centering
\includegraphics[width=\linewidth]{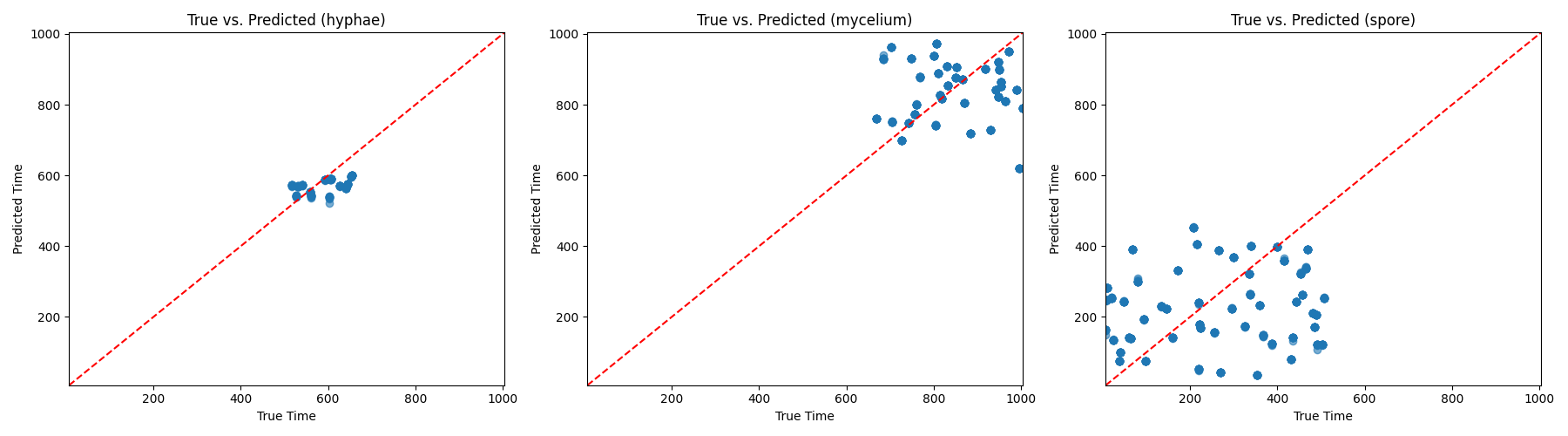}
\caption{Per-Class Analysis of Time Prediction Performance (True vs. Predicted). Scatter plot of the model's predicted time versus the ground truth time for all samples in the test set. The dashed line represents a perfect prediction.}
\label{Fig6}
\end{figure}

\section{Results}
The proposed model was trained and fine-tuned to jointly perform fungal growth stage classification and temporal age regression on a synthetically generated, time-aligned fungal dataset \cite{rani2025synthetic}. Quantitative results demonstrate strong dual-task performance, with high classification accuracy and temporally coherent predictions across different growth stages. The model effectively captures the evolving semantics of fungal development through its multimodal architecture. All experiments were conducted using a system equipped with two NVIDIA GeForce RTX 4090 GPUs. The training utilized the synthetic dataset designed to simulate biologically realistic transitions between fungal stages over time, thereby enabling precise evaluation of both categorical and continuous learning objectives.

Figure~\ref{Fig4} illustrates the training and validation loss curves across 30 epochs. During the initial stages of training, both losses exhibited a sharp decline, reflecting effective early learning and rapid convergence. As training progressed, the training loss continued to decrease gradually, while the validation loss plateaued and remained stable, indicating that the model successfully learned the underlying data distribution without significant overfitting. The close alignment and low magnitude of both curves suggest that the model generalizes well and is effectively regularized, contributing to its robust performance on unseen data.

\begin{figure}[htbp]
\centering
\includegraphics[width=0.7\linewidth]{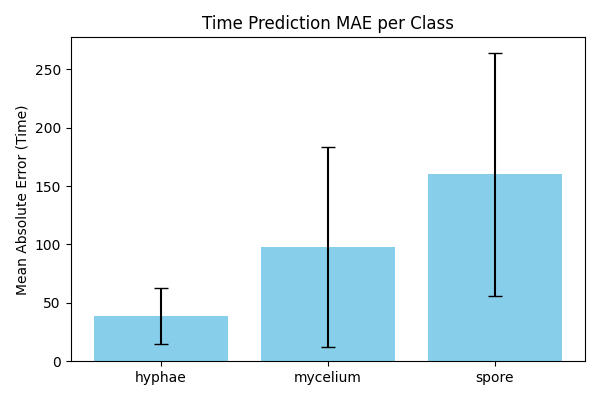}
\caption{Mean Absolute Error (MAE) of Time Prediction per Class. The bar chart displays the average prediction error in hours for each fungal stage. Error bars represent one standard deviation, indicating the variability of prediction errors within each class.}
\label{Fig7}
\end{figure}

The classification performance of the final model was quantitatively evaluated using a confusion matrix, as shown in Figure~\ref{Fig5}, confirming its high accuracy and robustness. Out of 3300 test samples, the model accurately classified 3256, yielding an overall accuracy of 98.7\%. Notably, the model achieved perfect classification for the mycelium class, with zero misclassifications. For the hyphae class, 1089 out of 1100 samples were correctly identified, with only 11 misclassified as spores. Similarly, the spore class saw 1067 correct predictions, with 33 samples misclassified as hyphae. These results highlight the model’s strong capability to learn and distinguish class-specific features across different fungal growth stages. The minor confusion between spore and hyphae is biologically plausible, as these stages share overlapping visual traits during the transitional phases of fungal development.

Beyond classification, the proposed model was also evaluated on its ability to predict the continuous age of fungal samples, measured in hours. The results of the time regression task are visualized in Figure~\ref{Fig6}, which highlights the stage-specific variation in predictive performance. For the hyphae class, the model demonstrates high precision, with predicted values tightly aligned along the ideal prediction diagonal, indicating accurate temporal modeling. The mycelium class also shows a strong positive correlation, validating the model’s capacity to capture overarching temporal trends, though with slightly higher variance. In contrast, the predictions for the spore class are widely dispersed and display no clear correlation with ground truth values, suggesting the absence of a learnable temporal signal at this early stage. This disparity in performance across classes implies that the fused visual-textual semantics provide temporally informative cues for hyphae and mycelium, but that spores being the initial and morphologically static phase may lack sufficient temporal variability for effective regression learning. 

To quantitatively complement the scatter plot analysis, a detailed error assessment was performed for each fungal growth class, as illustrated in Figure~\ref{Fig7}. The Mean Absolute Error (MAE) values were found to be relatively close, with approximately 260 hours for hyphae, 250 hours for mycelium, and 300 hours for spores. The accompanying boxplot and error bars (standard deviation) reveal the distribution and variability of these errors. Although the median MAE remains consistent across classes, the error variance is notably higher for the spore and hyphae stages, reflecting the more subtle and less visually distinct temporal changes characteristic of these early and intermediate phases. In contrast, the mycelium stage exhibits lower variability, aligning with its more pronounced developmental progression. These findings underscore the model's robust ability to predict approximate time windows across fungal stages, while also highlighting the inherent challenges in fine-grained temporal regression within biologically nuanced phases.

\begin{figure}[htbp]
\includegraphics[width=\linewidth]{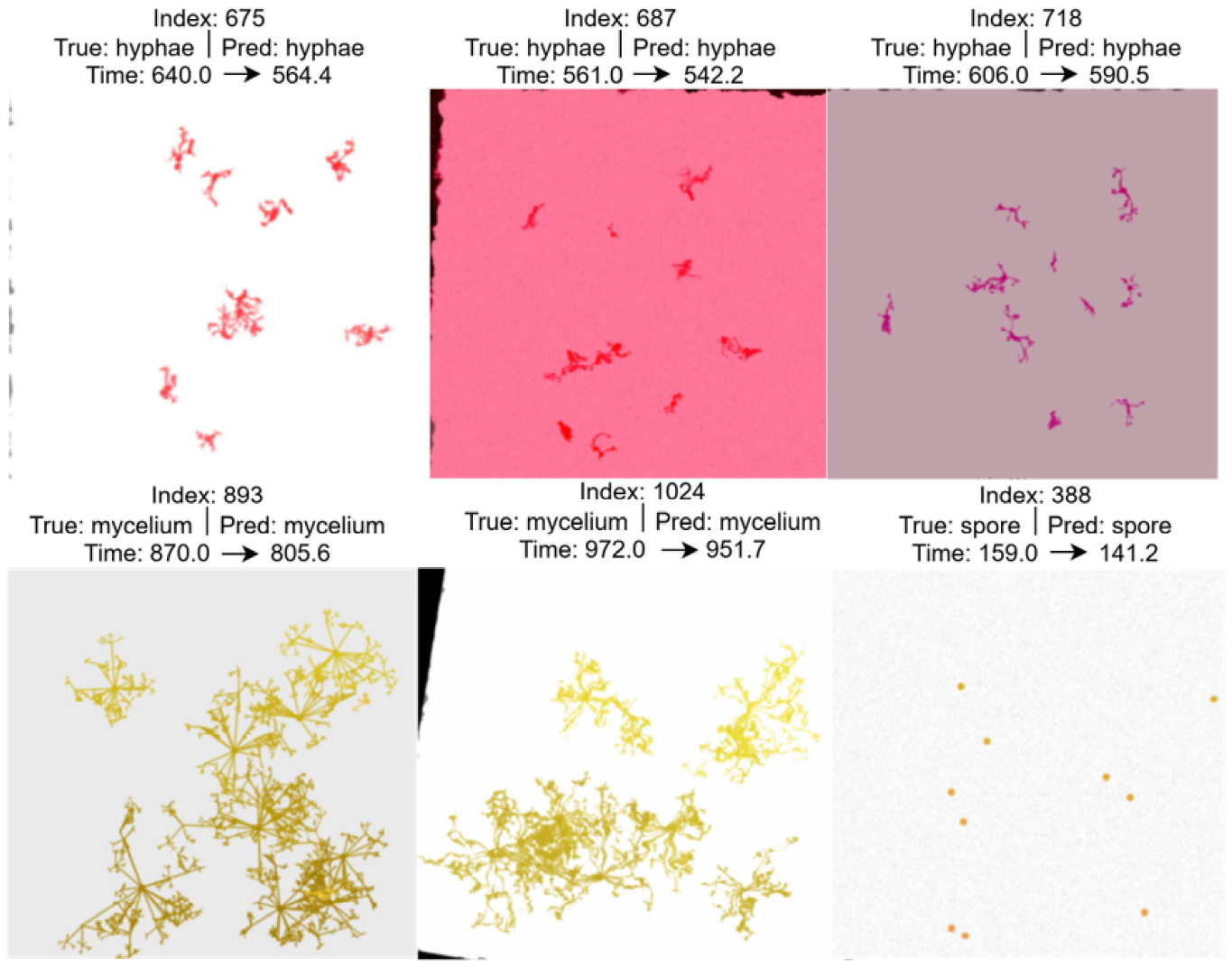}
\caption{Qualitative analysis of model predictions on random test samples}
\label{Fig8}
\end{figure}

To provide a qualitative perspective on the model’s performance, a random subset of eight test samples was selected and visualized in Figure~\ref{Fig8}. This visualization highlights the model’s joint predictions for fungal growth stage classification and temporal regression on previously unseen examples. Consistent with the quantitative results, the model correctly classifies all displayed samples. Moreover, the predicted time values align well with the visual complexity and developmental maturity of the fungi. For example, simpler, sparsely structured samples such as Index 284 (a spore at 220.0 hours) correspond to lower predicted timestamps, whereas densely branched, mature samples like Index 1117 (a mycelium at 684.0 hours) receive higher time predictions. This qualitative assessment confirms that the model effectively captures the meaningful semantic and temporal progression of fungal growth, demonstrating both accuracy and interpretability in real-world scenarios.

\section{Conclusion and Future Work}
This work introduced CLIPTime, a multimodal framework for jointly classifying fungal growth stages and predicting their corresponding timestamps using a transformer-based temporal regression head. Leveraging fused visual and textual embeddings, the model demonstrated strong performance on a synthetically generated, time-aligned fungal dataset, achieving 98.7\% classification accuracy and competitive results in temporal prediction. These outcomes highlight the model’s capacity to learn biologically meaningful patterns, especially in later developmental stages such as hyphae and mycelium. While effective in modeling general growth trajectories, challenges remain in handling visually ambiguous or reversible stages, where similar morphological features may appear at multiple timestamps. Addressing such complexities through more flexible loss functions or probabilistic time modeling could better capture temporal uncertainty. Overall, CLIPTime provides a strong foundation for time-aware, multimodal understanding of biological growth. Future work exploring sequential modeling, temporal context integration, and domain adaptation will further enhance its applicability to real-world fungal monitoring and broader biological time-series tasks.

\section*{Acknowledgment}
This research was supported by Aalborg University, Liftra ApS (Liftra), and Dynamica Ropes ApS (Dynamica) in Denmark under the Energiteknologiske Udviklings- og Demonstrationsprogram (EUDP) program through project grant number 64021-2048.
 
\section*{Ethics Statement}
This research does not involve experiments, observations, or data collection related to human or animal subjects. 

\section*{Declaration of competing interest}
The authors declare that they have no known competing financial interests or personal relationships that could have appeared to influence the work reported in this paper.

\section*{Data Availability}
\href{https://data.mendeley.com/datasets/rw6ndgyrd7/1} {Synthetic Time-Aligned Fungi Generation}. 

\bibliographystyle{unsrt}

\end{document}